# Title page

# AI-Based Fully Automatic Analysis of Retinal Vascular Morphology in Pediatric High Myopia

*Yinzheng Zhao[1, †], Zhihao Zhao[2, †], Junjie Yang[2], Li Li[3], M. Ali Nasseri[1, *], Daniel Zapp[1]*

[1] Klinik und Poliklinik für Augenheilkunde, Klinikum rechts der Isar, Technische Universität München, Munich, Germany
[2] Faculty of Information Technology, Technische Universität München, Munich, Germany
[3] Beijing Children's Hospital, Children's National Medical Center, Capital Medical University, Beijing, China

[†] These authors contributed equally in this work.

**\* Corresponding Author:**

Prof. M. Ali Nasseri

Ophthalmology Department of Klinikum rechts der Isar, Technische Universität München, Munich, Germany, 81675.

E-mail: ali.nasseri@mri.tum.de

**Funding:** Bavarian Research Alliance (AZ-1503-21)

**ABSTRACT**

**Purpose** To investigate the changes in retinal vascular structures associated with various stages of myopia by designing automated software based on an artificial intelligence model.

**Methods** The study involved 1324 pediatric participants from the National Children's Medical Center in China, and 2366 high-quality retinal images and corresponding refractive parameters were obtained and analyzed. Spherical equivalent refraction (SER) degree was calculated. We proposed a data analysis model based on a combination of the Convolutional Neural Networks (CNN) model and the attention module to classify images, segment vascular structures, and measure vascular parameters, such as main angle (MA), branching angle (BA), bifurcation edge angle (BEA) and bifurcation edge coefficient (BEC). One-way ANOVA compared parameter measurements between the normal fundus, low myopia, moderate myopia, and high myopia groups.

**Results** The mean age was 9.85±2.60 years, with an average SER of -1.49±3.16D in the right eye and -1.48±3.13D in the left eye. There were 279 (12.38%) images in the normal group and 384 (16.23%) images in the high myopia group. Compared with normal fundus, the MA of fundus vessels in different myopic refractive groups was significantly reduced (P = 0.006, P = 0.004, P = 0.019, respectively), and the performance of the venous system was particularly obvious (P<0.001). At the same time, the BEC decreased disproportionately (P<0.001). Further analysis of fundus vascular parameters at different degrees of myopia showed that there were also significant differences in BA and branching coefficient (BC). The arterial BA value of the fundus vessel in the high myopia group was lower than that of other groups (P = 0.032, 95% confidence interval [CI], 0.22–4.86), while the venous BA values increased (P = 0.026). The BEC values of high myopia were higher than those of low and moderate myopia groups. When the loss function of our data classification model converged to 0.09, the model accuracy reached 94.19%.

**Conclusion**

The progression of myopia is associated with a series of quantitative retinal vascular parameters, particularly the vascular angles. As the degree of myopia increases, the diversity of vascular characteristics represented by these parameters also increases.

**Introduction**

The alteration in the geometric shape of retinal vascular networks is an important indicator of vascular damage and is associated with various eye conditions,[1-6] including diabetic retinopathy,[5] hypertension-related retinal disorders,[7] pathological myopia[8] and glaucoma.[9] Additionally, the retinal vascular network can be directly observed in vivo, making it a subject of extensive attention and research aiming to identify predictive factors for various eye diseases.[10-13] The measurement of vascular network changes allows for the quantitative assessment of vascular pathological alterations.[11, 14, 15] This quantification, including vascular features such as blood vessel angles, vessel diameter, and vessel curvature, can enhance the clinical accuracy of related eye disease diagnosis by providing high precision. It also promotes the use of retinal vascular characteristics as a new biomarker in clinical diagnostic assistance.

The global prevalence of high myopia is increasing rapidly, and the risk of vision impairment or even loss due to pathological retinal changes associated with high myopia is significantly rising.[16-21] Retinal detachment associated with high myopia, along with other eye conditions such as myopic maculopathy and glaucoma,[22-24] not only increases the risk of outdoor accidents for individuals but can also potentially lead to blindness in severe cases.[25, 26] From an objective perspective of changes in the eye's structure, as the eye's axial length increases, there is enhanced tension on the posterior part of the eyeball.[27-31] In high myopic eyes, the axial length is longer, leading to greater tension, which makes the retinal vessel morphology more prone to changes.[22, 30, 32, 33] Recently, some researchers have also indicated a significant reduction in vessel diameter,[34] and a decrease in vessel fractal dimension in high myopia.[35] Additionally, Lim and his colleagues have pointed out a correlation between the degree of myopia and a decrease in retinal vessel angles.[36]

The complexity of retinal images, the subtle changes in vascular morphology, and the detailed measurement of vascular parameters all require advanced analytical techniques to effectively process large datasets. To address this, we have established a deep learning-based system for the assessment of retinal vascular parameters. A new indicator, main vessel angle, for evaluating the angle between retinal arteriovenous vessels is proposed. Through computer network analysis, various parameters of retinal vessels are calculated. This approach not only improves the accuracy of feature extraction and provides detailed and objective quantitative results but also reduces the inherent subjectivity of manual analysis.

## Methods

### Study Participants

Data were collected from 1324 children aged 1 to 17 years old. A total of 2366 images were included in this study. The Ethics Committee obtained approval from the Institutional Review Board at Beijing Children's Hospital, Capital Medical University. In accordance with the Helsinki Declaration, all participants provided informed consent before participating in the study. Since all participants were minors, professionals explained the purpose of the study, the data information, and the societal impact of the research to the guardians before data collection. After obtaining consent from the guardians, eye-related examinations and data collection were conducted.

### Data Collection

Each research subject underwent a comprehensive eye examination, including visual acuity testing, anterior segment examination, and fundus assessment. Non-cycloplegic refraction was conducted using an automatic refractometer (Topcon KR-8800, Japan). Digital fundus photography using a 45° retinal camera to capture fundus images centered on the macula (Topcon TRC-NW400, Japan). According to the "Expert Consensus on the Prevention and Control of High Myopia (2023)," the degree of myopia in the subjects was categorized into low, moderate, and high myopia (low myopia SER$\leqslant$-0.5 to >-3.00 diopter [D]; moderate myopia SER$\leqslant$-3.00D to >-6.00D; high myopia SER$\leqslant$-6.00D). Based on the SER of each eye of the subjects, their corresponding retinal fundus images were divided into four groups: the normal retinal fundus image group, the low myopia group, the moderate myopia group, and the high myopia group.

**Model System and Overview**

Currently, most studies investigate the relationship between myopia and retinal vasculature through causal inference methods. However, our research aims to identify biomarkers related to the development of high myopia, for which we constructed an AI classification model based on the Transformer architecture. The Transformer technology, a deep learning model based on the self-attention mechanism, can interact with the global information from different positions of the input data without being limited by the local window size. This allows the model to better understand the contextual information of the input data, thereby improving classification accuracy. The trained model achieved an accuracy of 94.19% after training. To understand the features learned by the model, we performed a visual analysis of the model's attention mechanism. The generated heatmap showed that the regions the model focused on when processing samples from different myopia groups were highly consistent with retinal vascular regions (**Fig. 1.**).

The second column might represent an initial detection heatmap, highlighting areas of interest or abnormalities. The color coding likely ranges from blue (low interest) to red or yellow (high interest), indicating regions that may contain pathological features. The third and fourth columns display the results of further analysis. These heatmaps are more detailed, with the focus areas gradually narrowing, indicating pathological conditions in the retinal images. The color intensity shifts from green to yellow, with the yellow gradient possibly indicating the most significant pathological areas. The reduced areas shown in bright yellow indicate greater confidence in the changes potentially related to the disease. Our algorithm and analysis process demonstrate high selectivity for pathological features, with the identified areas significantly overlapping with retinal vascular regions. Based on grouping conditions, it can be inferred that the severity of myopia is closely related to the morphology of retinal vasculature. This finding has prompted further attention and analysis of the retinal vascular structure.

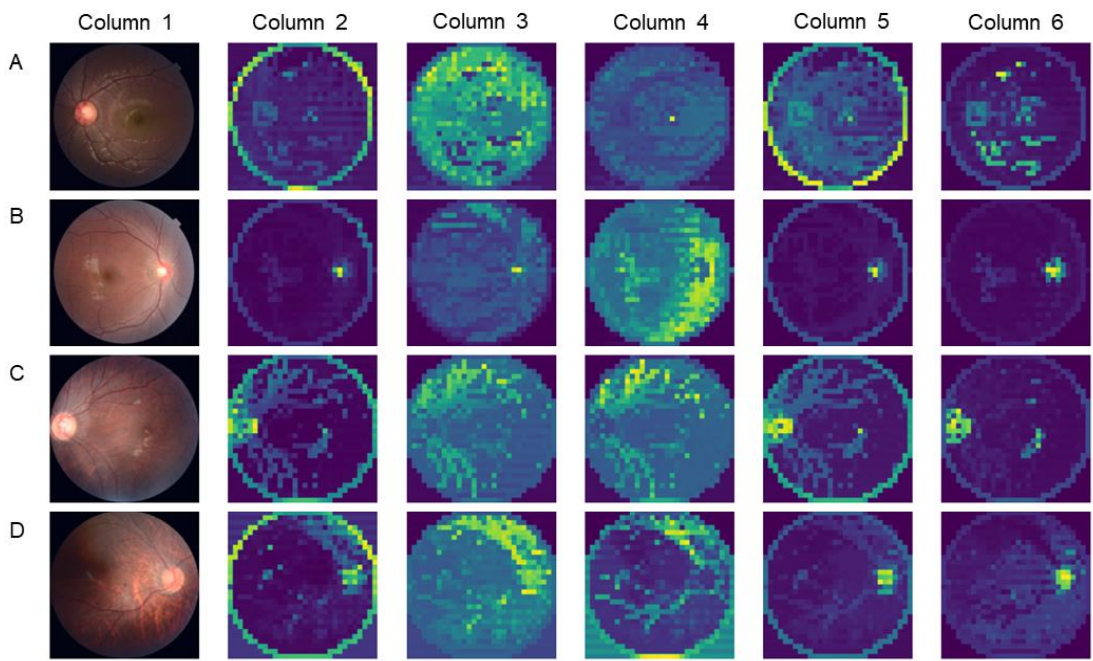

**Fig. 1. Different refractive stages of retinal image feature extraction maps.** Heatmaps were generated using the Attention module of the Transformer model. **A** represents original retinal images and various subsequent heatmaps for normal individuals, while **B** represents representative images for the low myopia group, and **C**, **D** represent the moderate and high myopia groups, respectively. The first column of images consisted of original high-resolution retinal images for four groups with different refractive statuses, including the retina, optic disc, macula, and posterior pole. Columns 2 through 6 displayed heatmaps generated by the Attention module, capturing the regions of interest.

Consequently, we designed a vascular quantification analysis system, which consists of three parts. First (**Fig. 2A**), high-resolution fundus images are obtained through image quality assessment. In the experiments, based on expert consensus, we use SER values to classify retinal images, roughly dividing them into four categories: normal retina, low myopia, moderate myopia, and high myopia. Second (**Fig. 2B**), to better segment the arteries and veins of the vascular structure, our model adopts U-Net with topological structure constraints, which can effectively handle interrupted parts in vascular images and improve the segmentation of small vessels.[37] After segmenting arteries and veins, we perform skeletonization of the vascular structure while detecting branching points and endpoints. Third (**Fig. 2C**), we apply Strahler order[38] to assign three important landmarks. Crossings occur where two vessels intersect, while branching and bifurcation points occur where a single vessel splits into two. Branching points are where a smaller vessel branches off from the main vessel, with differing vascular orders in the daughter vessels. Bifurcation points are where a vessel divides into two similar vessels, with the same vascular order in the daughter vessels. After that, the main stem, bifurcation points, and branching points of the blood vessels are determined, and then the corresponding blood vessel parameters are calculated.

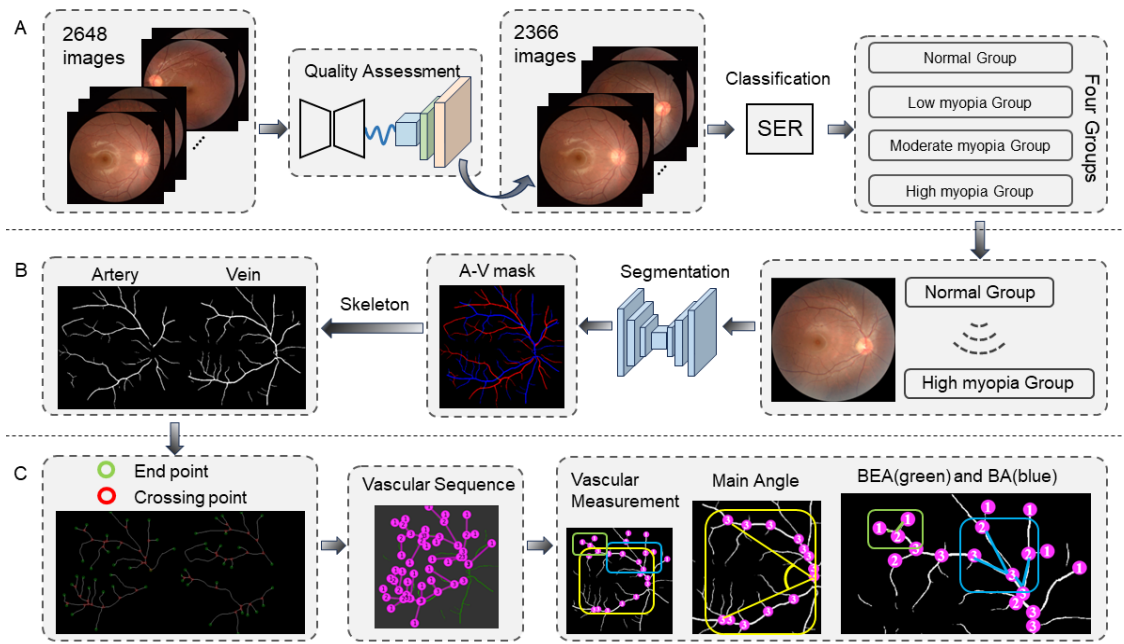

**Fig. 2**. **Model workflow diagram**. **A** primarily involves data quality analysis and classification. **B** carries out image segmentation and arteriovenous skeletonization based on the results from **A**. Building upon **B**, the model further determines vascular order (**C**) and measurement parameters.

**Parameters Measurement**

In our academic report, the restricted measurement of the retinal branch vascular area was defined as ranging from 0.5 to 2.0-disc diameters from the optic disc edge, based on the vascular structure features extracted from retinal images. The BA and BC related to the branch angle were obtained in this area. The largest two arteries and two veins in terms of vessel diameter were selected as the basis for measuring the MA. By determining the vascular order, we could initially locate the main vessel of the artery or vein with the highest vascular order. The model then obtained the value of MA by connecting the corresponding points to the center of the optic disc (**Fig. 3**). Based on the rules for vascular order and branching points (also bifurcation points), as elucidated in the Methods section, an analysis of parameters including MA, BA, BC, BEA, and BEC was conducted.[39, 40] Among these, the measurement methods for BA and BEA were illustrated by the blue and green boxes in **Fig. 3**. The measurements of BC and BEC were both related to the diameter of the parent segment vessel and the diameter of the daughter segment vessel.[39]

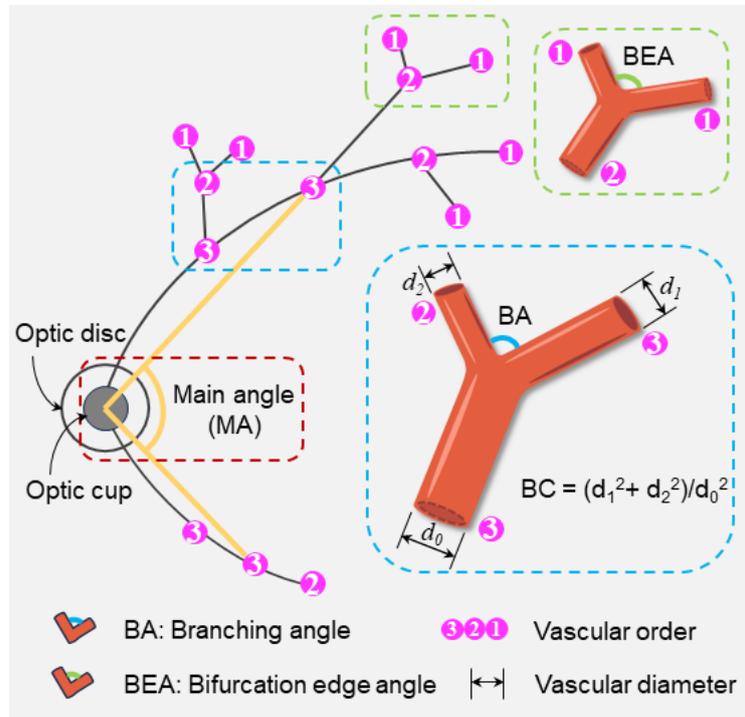

**Fig. 3. Diagram illustrating vascular measurement parameters.** Arteriole (or venule) vessels corresponding to the temporal side of the optic disc in the image are used as reference points. Numbers represent vascular orders. The angle formed by the last vessel with the highest vascular order on both upper and lower vessels intersecting at the optic cup is defined as MA. BA is formed by vessels with different vascular orders, while BEA is formed by vessels with the same vascular order. The calculation method for the branching coefficient is related to the diameter of the branching vessel and the diameter of the parent vessel.

**Data Analysis**

Python 3.10.13, SPSS V.27.0 (SPSS Inc., Chicago, IL), and Origin (OriginLab, Northampton, US-MA) softwires were employed for data model training, statistical analysis, and graphical representation. Main vessel angles, branching angles, bifurcation edge angles, branching coefficient, and bifurcation edge coefficient were all quantitative data. Descriptive statistics were presented in the form of the mean and standard deviation (SD) of the data. One-way analysis of variance (ANOVA) was used to compare groups of different refractive error statuses.

**Results**

This study included 1324 retinal photographs from both eyes of children, among which there were 713 boys and 611 girls. The mean age was 9.85±2.60 years, with an average SER of -1.49±3.16 D in the right eye and -1.48±3.13 D in the left eye. A total of 2648 retinal fundus images were included. Due to low image quality, poor resolution, and partial image loss, 282 images were excluded during screening, resulting in a final study cohort of 2366 images (**Table 1.**). There were 279 images in the normal retinal fundus image group, 1193 images in the low myopia group, 510 images in the moderate myopia group, and 384 images in the high myopia group.

**Table 1** summarized the data characteristics and vascular branching parameters of retinal arteries and veins in different refractive status groups. There was a statistically significant correlation between individual age and the severity of myopia ($P<0.001$). With increasing age, there was a trend toward progressively deepening myopia. **Fig. 4** showed the performance of vascular parameters in the arterial (**Fig. 4A, 4B, and 4C**) and venous systems (**Fig. 4D, 4E, and 4F**) across four groups with different degrees of myopia. The differences in the structure and function of arteries and veins in the eye result in distinct hemodynamic and blood pressure characteristics. Analyzing the vascular parameters of arteries and veins separately allows for a more accurate assessment and understanding of pathological changes. Comparing the measurement results of retinal artery and vein parameters in different groups, significant statistical correlations (**Fig. 4A**, $P<0.05$) were found between the MA values corresponding to different myopia stages in the artery parameters. A similar phenomenon was also observed in the vein system parameters (**Fig. 4D**, P<0.001). In addition, the P-values for BEC in both arteries and veins were less than 0.001, indicating significant differences in BEC across different myopia stages. Compared to normal retinal fundus images, retinal fundus images with myopia showed a significant decrease in BEC values, with the decrease being more pronounced in veins than in arteries. The remaining indicators, including BAC and BEA, had not shown statistically significant changes. Further analysis would be conducted to examine the parameter changes among different myopia stages.

    **Table 2** described comparisons between different groups (normal and low myopia, normal and moderate myopia, normal and high myopia) across various parameters. Each parameter included the mean difference (MD) with a 95% confidence interval (CI) and a P value indicating the statistical significance of the difference. During the transition from normal retinal fundus images to myopic retinal fundus images, there was a significant decrease in the main angle of the arteries (P = 0.006, P = 0.004, P = 0.019). In the low, moderate, and high myopia groups, the angle difference decreases, but the angle change tended to stabilize. Unlike the arterial vessel angle, the main angle of the veins continued to decrease (P = 0.002, P<0.001, P<0.001), with a change magnitude far greater than that of the arterial vessels. Similar to MA, BEC values also showed significant relevance with myopia status in the comparison of arterial and venous systems (**Fig. 4C and 4F**, P<0.05), negating random variation in differences. However, BEA values did not consistently show significant differences between normal

individuals and myopia patients in retinal fundus images. In the comparison between the normal group and the high myopia group, differences were observed in the arterial BA values (**Fig. 4B**), indicating that significant changes in branching angles mainly occur at higher levels of myopia severity.

The retinal vascular morphology in normal individuals differed significantly from that in myopia patients, with significant correlations observed with the progression and severity of myopia-related changes. Further analysis of changes in retinal vascular parameters in the arterial and venous systems across different myopia stages revealed that inter-group differences tended to stabilize (**Table 3**). In the arterial system, both BC and BEC decreased gradually as the severity of myopia deepened (P<0.05). A similar trend parallel to this was observed in the venous system, although the differences were not significant. The changes in the main angle from low myopia to high myopia had a significant impact on venous structure (P = 0.003, P = 0.036). Compared with moderate myopia, there was a statistical correlation between BA values in patients with high myopia (**Fig. 4B and 4E**). The values of arterial branches decreased (P = 0.032) and the angle of venous branches increased (P = 0.026). However, this change is not shown in the normal fundus. Comparing the comparison between the normal group and the myopia groups, there was a lack of significant changes in structural parameters in the analysis of different degrees of myopia groups, while the differences were mainly concentrated in the changes in high myopia parameters.

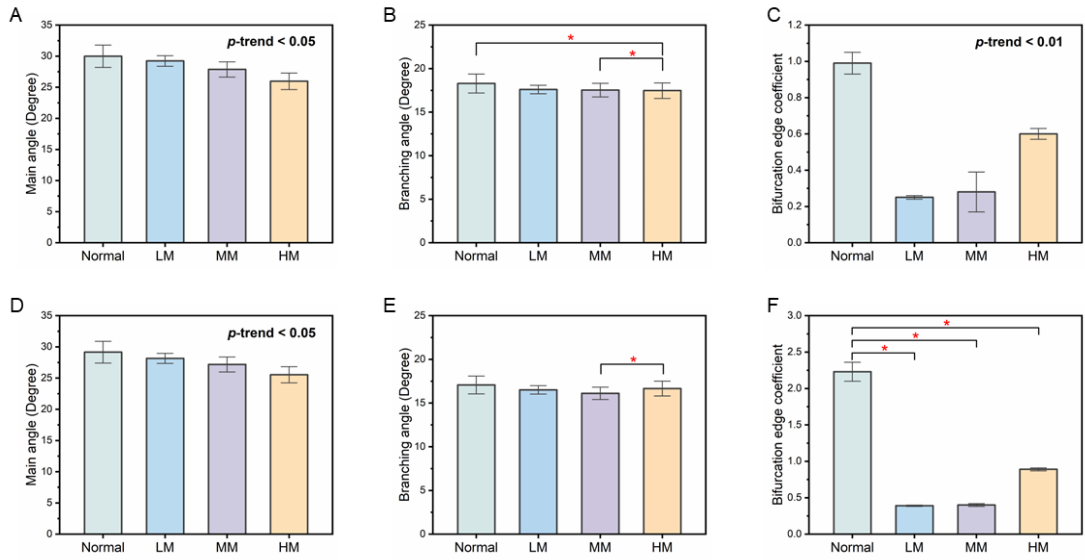

**Fig. 4.** The standard deviation bar chart displayed the differences in vascular parameters as arterial main angle (A), arterial branching angle (B), arterial bifurcation edge coefficient (C), venous main angle (D), venous branching angle (E), and venous bifurcation edge coefficient (F) between the four groups: normal, low myopia (LM), moderate myopia (MM), and high myopia (HM). Special markers (✱) were considered to demonstrate a statistical difference between two sets of data.

**Table 1.** Data characteristics and retinal vascular measurement parameters by refractive status

| Characteristics | Normal | Low myopia | Moderate myopia | High myopia | P value |
|---|---|---|---|---|---|
| Sample size | 119 | 646 | 303 | 256 | |
| Fundus images | 279 | 1193 | 510 | 384 | |
| Age | 8.68±0.224 | 9.79±0.082 | 11.35±0.130 | 11.42±0.401 | **<0.001** |
| Boys | 0.10±0.04 | -1.70±0.04 | -4.06±0.06 | -6.73±0.13 | |
| Girls | 0.18±0.07 | -1.66±0.04 | -4.03±0.07 | -6.88±0.19 | |
| Vascular measurement parameters | | | | | |
| Artery (95% CI) | | | | | |
| Main angle | 87.55 (84.02 to 91.08) | 82.12 (80.46 to 83.78) | 82.44 (79.01 to 83.86) | 82.43 (79.83 to 85.03) | **0.023** |
| Branching angle | 36.28 (34.12 to 38.44) | 34.86 (33.90 to 35.83) | 35.88 (34.35 to 37.40) | 33.34 (31.59 to 35.09) | 0.092 |
| Branching coefficient | 1.67 (1.37 to 1.96) | 1.31 (1.28 to 1.34) | 1.55 (1.16 to 1.94) | 1.41 (1.33 to 1.50) | **0.048** |
| Bifurcation edge angle | 34.97 (33.34 to 36.60) | 34.91 (34.21 to 35.61) | 34.68 (33.66 to 35.71) | 33.73 (32.43 to 35.02) | 0.434 |
| Bifurcation edge coefficient | 0.92 (0.79 to 1.04) | 0.72 (0.70 to 0.74) | 0.70 (0.68 to 0.73) | 0.78 (0.72 to 0.84) | **<0.001** |
| Vein (95% CI) | | | | | |
| Main angle | 84.94 (81.50 to 88.37) | 79.05 (77.45 to 80.65) | 78.07 (75.70 to 80.43) | 74.31 (71.75 to 76.87) | **<0.001** |
| Branching angle | 36.19 (34.18 to 38.20) | 36.44 (35.50 to 37.38) | 35.24 (33.84 to 36.65) | 37.71 (36.04 to 39.39) | 0.175 |
| Branching coefficient | 1.36 (1.20 to 1.52) | 2.35 (0.50 to 4.20) | 1.34 (1.04 to 1.65) | 1.31 (1.23 to 1.39) | 0.769 |
| Bifurcation edge angle | 34.08 (32.74 to 35.42) | 35.22 (34.64 to 35.81) | 34.65 (33.82 to 35.48) | 34.05 (32.90 to 35.20) | 0.147 |
| Bifurcation edge coefficient | 0.94 (0.67 to 1.21) | 0.60 (0.58 to 0.62) | 0.58 (0.55 to 0.61) | 0.65 (0.57 to 0.73) | **<0.001** |

0

**Table 2.** Comparison of retinal arterial and venous blood vessel measurements between normal group and different refractive status groups

| Comparison between groups | Main angle | | Branching angle | | Branching coefficient | | Bifurcation edge angle | | Bifurcation edge coefficient | |
|---|---|---|---|---|---|---|---|---|---|---|
| | MD (95% CI) | P value | MD (95% CI) | P value | MD (95% CI) | P value | MD (95% CI) | P value | MD (95% CI) | P value |
| Artery | | | | | | | | | | |
| Normal and Low myopia | 5.43 (1.59 to 9.26) | **0.006** | 1.42 (0.22 to 3.61) | 0.215 | 0.35 (0.19 to 0.51) | **<0.001** | 0.62 (-1.59 to 1.72) | 0.941 | 0.19 (0.13 to 0.26) | **<0.001** |
| Normal and Moderate myopia | 6.11 (1.92 to 10.29) | **0.004** | 0.40 (0.19 to 3.01) | 0.761 | 0.11 (0.05 to 0.68) | 0.696 | 0.28 (-1.55 to 2.11) | 0.762 | 0.21 (0.12 to 0.31) | **<0.001** |
| Normal and High myopia | 5.12 (0.84 to 9.40) | **0.019** | 2.94 (0.19 to 5.69) | **0.036** | 0.25 (0.02 to 0.53) | 0.067 | 1.24 (-0.82 to 3.29) | 0.236 | 0.13 (0.00 to 0.26) | **0.044** |
| Vein | | | | | | | | | | |
| Normal and Low myopia | 5.89 (2.19 to 9.59) | **0.002** | 0.25 (-2.42 to 1.92) | 0.82 | -0.99 (-4.82 to 2.83) | 0.612 | -0.68 (-1.97 to 0.61) | 0.301 | 0.34 (0.20 to 0.47) | **<0.001** |
| Normal and Moderate myopia | 6.87 (2.79 to 10.94) | **<0.001** | 0.94 (-1.46 to 3.35) | 0.441 | 0.02 (-0.41 to 0.45) | 0.935 | -0.16 (-1.54 to 1.23) | 0.824 | 0.36 (0.16 to 0.56) | **<0.001** |
| Normal and High myopia | 10.63 (6.44 to 14.82) | **<0.001** | 1.73 (-4.32 to 0.86) | 0.19 | 0.05 (-0.12 to 0.21) | 0.561 | 0.43 (-1.23 to 2.09) | 0.611 | 0.29 (0.04 to 0.53) | **0.022** |

**Table 3.** Comparison of retinal arteriovenous vessel measurements in low, moderate and high myopia

| Comparison between groups | Main angle | | Branching angle | | Branching coefficient | | Bifurcation edge angle | | Bifurcation edge coefficient | |
|---|---|---|---|---|---|---|---|---|---|---|
| | MD (95% CI) | P value | MD (95% CI) | P value | MD (95% CI) | P value | MD (95% CI) | P value | MD (95% CI) | P value |
| Artery | | | | | | | | | | |
| Low myopia and Moderate myopia | 0.68 (-2.31 to 3.67) | 0.654 | -1.01 (-2.78 to 0.76) | 0.263 | -0.24 (-0.50 to 0.02) | 0.067 | 0.22 (-1.03 to 1.47) | 0.726 | 0.02 (-0.01 to 0.05) | 0.253 |
| Low myopia and High myopia | -0.31 (-3.59 to 2.97) | 0.853 | 1.52 (-0.44 to 3.49) | 0.128 | -0.10 (-0.17 to -0.03) | **0.008** | 1.18 (-0.26 to 2.62) | 0.109 | -0.06 (-0.11 to -0.02) | **0.007** |
| Moderate myopia and High myopia | -0.99 (-4.58 to 2.60) | 0.588 | 2.54 (0.22 to 4.86) | **0.032** | 0.14 (-0.31 to 0.59) | 0.54 | 0.96 (-0.67 to 2.58) | 0.248 | -0.08 (-0.14 to -0.02) | **0.008** |
| Vein | | | | | | | | | | |
| Low myopia and Moderate myopia | 0.98 (-1.91 to 3.87) | 0.506 | 1.19 (-0.51 to 2.90) | 0.168 | 1.01 (-1.83 to 3.84) | 0.486 | 0.57 (-0.47 to 1.62) | 0.282 | 0.02 (-0.02 to 0.06) | 0.352 |
| Low myopia and High myopia | 4.74 (1.57 to 7.91) | **0.003** | -1.27 (-3.18 to 0.63) | 0.19 | 1.04 (-2.23 to 4.31) | 0.533 | 1.18 (-0.04 to 2.39) | **0.058** | -0.05 (-0.11 to 0.01) | 0.096 |
| Moderate myopia and High myopia | 3.76 (0.25 to 7.27) | **0.036** | -2.47 (-4.64 to -.30) | **0.026** | 0.03 (-0.33 to 0.39) | 0.867 | 0.60 (-0.78 to 1.98) | 0.392 | -0.07 (-0.15 to 0.01) | 0.082 |

**Discussion**

In the realm of ophthalmic research, the correlation between myopia progression and its associated retinal vascular changes represents a critical area of study,[41-43] offering insights into the pathophysiological mechanisms underlying myopia and its varying degrees of severity. The changes in retinal vascular diameter and other geometric features, including branching angles and bifurcation angles, have been recognized as significant indicators of microvascular alterations.[14, 39] Our AI model focuses intensely on the retinal vascular structure in myopic refractive errors, considering the anatomical basis of myopia formation, which includes ocular expansion,[44] scleral stretching,[45] and anomalous ocular morphology due to adjacent tissue compression.[46] These factors, along with structural changes in retinal tissue and potential neovascularization, interact to result in abnormal morphology and distribution of retinal vessels.[47] We believe that the peripheral retinal vascular structure loss in myopic patients compromises the optimal structure of the retinal vasculature system for blood transport.

By comparing the retinal vascular structures of normal and myopic fundus images, the study identified a notable aspect. Within the arterial parameters range, there is a significant statistical correlation between the main angles (MA) and different stages of myopia, as well as a similar trend observed within the venous system parameters. Further analysis showed that as myopia progresses from low to high degrees, the reduction in MA values is more pronounced in the venous system, indicating that the venous main vessel structures undergo greater changes as myopia progressively

30  worsens, and the thinner venous walls are more susceptible to morphological changes

31  due to deformation of the eyeball. Significant and profound changes occurred in the

32  BEC values within arterial and venous structures during the transition from normal

33  fundus to different stages of myopic fundus. It is particularly noteworthy that, compared

34  to arteries, there is a more apparent reduction in BEC values in veins, indicating that

35  compared to the retinal vascular diameters of a normal fundus, the vascular diameters

36  in the myopic fundus significantly decrease. The difference in the vessel angle of the

37  venous system as compared to the arterial system suggests a unique vascular

38  remodeling pattern in myopic eyes, which can reflect potential microvascular changes

39  associated with the development of myopia.

40  Further research into the morphological changes of fundus vessels at different

41  stages of myopia revealed that although the value of the peripheral vascular branching

42  factor significantly decreased during the transition from a normal fundus to low myopia,

43  there was a trend of increase when myopia progressed from low to high degrees. The

44  progression of myopia may affect hemodynamics by altering the internal structure of

45  the eye.[48] Hong et al. pointed out that the structural changes of the eyeball in myopia

46  affect parameters such as retinal blood vessel density.[49] The shear stress generated

47  at the branching points causes an adaptive response in the diameter of the vessels.[50,]

48  [51] This response may lead to a disproportionate increase in branch vessels relative to

49  the parent vessel, resulting in higher BEC values. The stress and remodeling produced

50  by the vascular system, through the expansion of marginal subsidiary vessels, ensure

51  sufficient blood perfusion to the peripheral retina. This change may be a compensatory response to the increased oxygen demand caused by retinal morphological changes due to high myopia.

    This study also emphasized that compared to non-high myopia, patients with high myopia had significantly decreased arterial branching angles and increased venous branching angles. We believe that these changes in the vascular structure of the fundus in high myopia may be related to the adaptive requirements of the retina. To accommodate the elongation of the eyeball structure and the potentially increased metabolic demand due to the increased retinal surface area,[52, 53] the arterial system BA at smaller values to promote more directed blood flow, thereby effectively covering the peripheral areas of the retina and ensuring sufficient tissue perfusion. At the same time, the increase in intraocular pressure and biochemical changes in the vascular walls caused by high myopia led to increased arterial stiffness and a reduction in vascular angles.[51] Smaller branching angles optimized the distribution of blood flow and sheared stress in highly myopic eyes to the maximum extent.[40] The increase in venous branching angles may reflect an optimized strategy for blood reflux in the retinas of patients with high myopia. Larger branching angles may help reduce the venous resistance caused by increased intraocular pressure due to axial elongation and improve blood flow. The change in venous branching angles may also be a compensatory response to the decrease in arterial branching angles. Arterial blood flow based on the reduced BA reaches deeper peripheral regions of the retina, and veins need to recover blood at larger angles effectively to return it to the central circulation.

The observed differences in the adaptation of arterial and venous branching angles in patients with high myopia emphasize the eye's complex response to the hemodynamic changes caused by the elongation of the myopic eye axis. These changes not only reflect the adaptive mechanisms to maintain ocular perfusion integrity but also highlight the importance of considering vascular factors in the management and treatment of myopia and its complications. Further research into these vascular adaptations could provide insights into new therapeutic targets for myopia and its related ocular changes.

This study has some limitations. The first significant limitation is the bias in image selection and image analysis algorithms. The study excluded 282 images due to factors such as poor image quality, low resolution, or partial image loss based on image quality assessment. This exclusion may introduce selection bias, potentially leading to a sample that does not fully represent the broader population under study. Additionally, reliance on image analysis by the model introduces the possibility of algorithmic bias, meaning the model may not be applicable to all image qualities. Secondly, considering the time efficiency of large-scale data collection and the acceptability of the procedure by participants, the study opted for non-cycloplegic automated refraction. However, this method may result in inaccurate measurements of refractive errors due to incomplete relaxation of the ciliary muscle. Future studies should ideally use cycloplegic refraction to ensure reliable measurement results. Finally, the cross-sectional nature of this study limits the ability to infer causality between retinal vascular changes and myopia progression. In future research, using longitudinal data to address these limitations, validate the findings, and enhance the generalizability of the results will be crucial.

**Conclusion**

In summary, the significant correlations observed between retinal vascular structures in normal and myopic populations, as well as the different vascular remodeling patterns found in myopic eyes, emphasize the complexity of myopia as a multifactorial condition. These insights not only enhance our understanding of the pathophysiological mechanisms of myopic eyes but also highlight the potential for new diagnostic and therapeutic approaches targeting the vascular morphology of myopic eyes. Further research is necessary to explore the longitudinal effects of these vascular changes on visual function and to assess the potential of intervention strategies aimed at mitigating the progression of myopia and its related ocular complications.

**Acknowledgements:** Not applicable.

**Data Availability:** The relevant data have been provided in the manuscript. Raw image datasets used and/or analyzed during the current study are available from the corresponding author upon reasonable request.

**Declarations**

• **Funding**: Supported by Bavarian Research Alliance (AZ-1503-21)

• **Competing interests**: The authors declare no competing interests.

• **Ethics approval and consent to participate**: This study adhered to the principles of the Declaration of Helsinki. The study protocol was reviewed and approved by the Ethics Committee of Beijing Children's Hospital (approval number [IEC-C-008-A08-V.05.1]. Signed informed consent was obtained from all parents.

- **Consent for publication**: Not applicable.
- **Author contribution**: Y.Z. is the first author. M.N. obtained funding. Y.Z. and Z.Z. designed the study. Y.Z., Z.Z., and J.Y. collected the data. Y.Z. and Z.Z. are involved in data cleaning, and data analysis. Y.Z. drafted the manuscript. M.N., D.Z., and L.L. contributed to the critical revision of the manuscript for important intellectual content and approved the final version of the manuscript. All authors have read and approved the final manuscript. M.N. is the Corresponding Author.